\documentclass{article}

\usepackage{microtype}
\usepackage{graphicx}
\usepackage{subfigure}
\usepackage{booktabs} 
\usepackage{hyperref}

\usepackage[accepted]{icml2025}

\usepackage{amsmath}
\usepackage{amssymb}
\usepackage{mathtools}
\usepackage{amsthm}

\usepackage[capitalize,noabbrev]{cleveref}

\theoremstyle{plain}

\theoremstyle{definition}

\theoremstyle{remark}

\usepackage[textsize=tiny]{todonotes}

\icmltitlerunning{Transfer Learning for Onboard Cloud Segmentation in Thermal EO: From Landsat to a CubeSat Constellation}

\begin{document}

\twocolumn[
\icmltitle{Transfer Learning for Onboard Cloud Segmentation in Thermal Earth Observation: From Landsat to a CubeSat Constellation}

\icmlsetsymbol{equal}{*}

\begin{icmlauthorlist}
\icmlauthor{Niklas Wölki}{orora}
\icmlauthor{Lukas Liesenhoff}{orora}
\icmlauthor{Christian Mollière}{orora}
\icmlauthor{Martin Langer}{orora}
\icmlauthor{Julia Gottfriedsen}{orora}
\icmlauthor{Martin Werner}{tum}
\end{icmlauthorlist}

\icmlaffiliation{orora}{OroraTech GmbH, Munich, Germany}
\icmlaffiliation{tum}{Technical University of Munich (TUM), Munich, Germany}

\icmlcorrespondingauthor{Niklas Wölki}{niklas.woelki@ororatech.com}
\icmlcorrespondingauthor{Lukas Liesenhoff}{lukas.liesenhoff@ororatech.com}

\icmlkeywords{Transfer Learning, Cloud Detection, Domain Adaptation, Computer Vision, CubeSat, Onboard processing}

\vskip 0.3in
]
\printAffiliationsAndNotice{}

\begin{abstract}
Onboard cloud segmentation is a critical yet underexplored task in thermal Earth observation (EO), particularly for CubeSat missions constrained by limited hardware and spectral information. CubeSats often rely on a single thermal band and lack sufficient labeled data, making conventional cloud masking techniques infeasible. This work addresses these challenges by applying transfer learning to thermal cloud segmentation for the FOREST-2 CubeSat, using a UNet with a lightweight MobileNet encoder. We pretrain the model on the public Landsat-7 Cloud Cover Assessment Dataset and fine-tune it with a small set of mission-specific samples in a joint-training setup, improving the macro F1 from 0.850 to 0.877 over FOREST-2-only baselines. We convert the model to a TensorRT engine and demonstrate full-image inference in under 5 seconds on an NVIDIA Jetson Nano. These results show that leveraging public datasets and lightweight architectures can enable accurate, efficient thermal-only cloud masking on-orbit, supporting real-time decision-making in data-limited EO missions.
\end{abstract}

\section{Introduction}
\label{sec:introduction}
In times of accelerating climate change, Earth observation (EO) satellites have become essential tools for monitoring natural hazards and supporting disaster response \cite{Denis2016-tj}. Thermal remote sensing, in particular, plays a critical role in applications such as wildfire detection \cite{wildfire_dima}, urban heat monitoring \cite{molliere2024subdaily}, and agricultural drought assessment \cite{andersonDrought}. Thermal EO requires frequent, high-resolution data to enable actionable insights, a combination that few traditional missions can offer. Although Landsat-7 ETM+ sensors provide a spatial resolution of 60m, they suffer from a low temporal resolution of 16 days \cite{landsat7-tir}. In contrast, instruments such as MODIS have a quasi-daily revisit frequency but a coarse thermal spatial resolution of 1 km \cite{modis_revisit}. New-generation CubeSat constellations, such as the OTC-P1 constellation of OroraTech GmbH, address this gap by delivering thermal imagery on a global scale with a revisit rate of twice daily. 

However, clouds remain a significant obstacle to EO. They obstruct ground visibility and significantly alter the measured signal, making an accurate cloud mask essential. Traditional cloud detection algorithms rely heavily on multispectral or visible imagery, often unavailable on resource-constrained CubeSats, especially for thermal-only imaging, where clouds are difficult to distinguish from other cold surface structures. Although there is ample work on multispectral methods, cloud segmentation with thermal bands remains largely underexplored. 

At the same time, machine learning methods, particularly deep learning, have demonstrated superior performance over traditional cloud masking approaches \cite{li_cloud_2022}. Despite their effectiveness, deep learning methods heavily rely on large, well-annotated datasets, which are often lacking in the early stages of CubeSat missions. Additionally, mission operators may not have fully calibrated the instruments in early phases.

We explore transfer learning using existing high-quality public datasets to address this challenge. Specifically, we leverage the Landsat-7 Cloud Cover Assessment Dataset \cite{l7-dataset} to improve cloud segmentation performance for the OTC-P1 constellation, consisting of eight FOREST-2 satellites. Following recent cross-mission cloud detection efforts, such as the adaptation from Landsat-8 to Proba-V by \citet{MateoGarca2020}, we evaluate cross-domain generalization and joint training strategies.

Our experiments demonstrate that incorporating Landsat labels significantly enhances thermal cloud segmentation on CubeSat imagery. We combine approximately 6,000 image crops from the Landsat-based Cloud Cover Assessment Dataset with around 500 labeled crops from the FOREST-2 mission. Compared to training solely on CubeSat data, joint training with Landsat yields more balanced predictions and improves the macro F1 score from 0.850 to 0.877 across a 6-fold spatial cross-validation. These results highlight the potential of pretraining models on global, geographically diverse datasets from established missions and fine-tuning them with a small set of target-domain samples, offering a promising strategy for early-stage CubeSat missions with limited labeled data.

\section{Data}
\label{sec:data}
We use two datasets in our experiments: a manually labeled thermal dataset from the FOREST-2 CubeSat mission, the predecessor of the OTC-P1 constellation, and the Landsat-7 Cloud Cover Assessment Dataset \cite{l7-dataset}.

The FOREST-2 dataset consists of 528 image crops (256 × 256 pixels) derived from 24 hand-labeled thermal scenes captured by the FOREST-2 CubeSat, where data is more frequent over wildfire-prone regions such as Australia and North America. The satellite employs two long-wave infrared and one mid-wave infrared band. This work focuses on the LWIR2 band (10.5-12.6 µm) as a single-band input to the models, as Landsat-7 employs a matching band.

We evaluated several candidate missions for the complementary dataset, including VIIRS, MODIS, and Sentinel-3, but identified Landsat-7 as the best fit. Its Band 6 offers a spectral response function (SRF) that closely matches the LWIR2 band of our sensor, making it particularly suitable for transfer learning. Additionally, Landsat-7 provides a spatial resolution of 60m in the thermal band, which we can reliably downsample to match the 200m resolution of FOREST-2. Finally, a well-established, high-quality, manually labeled cloud mask validation dataset is available for this mission. The dataset includes 6,000 image crops, which we extracted from 206 globally distributed scenes spanning diverse biomes. Given the close match in spectral response between Landsat-7 and FOREST-2, we choose to apply only spatial domain adaptation by bilinearly resampling the Landsat crops to a ground sampling distance (GSD) of 200m and 256 x 256 px size.

For the FOREST-2 dataset, we adopt a 6-fold cross-validation strategy with spatial blocking, following the approach proposed by \citet{kattenborn_spatially_2022}, to mitigate the effects of spatial autocorrelation and the limited dataset size. In contrast, for the large and globally representative Landsat-7 dataset, we choose to use a traditional train-test split.
For each experiment where the FOREST-2 dataset is part of the training, on every iteration, we use four folds for training, one fold for validation, and hold out the remaining fold for testing. We repeat this process across all six folds and report the final performance as the average of the six test scores. We aim to generate binary cloud masks, classifying each pixel as either cloud or clear sky.

\section{Methods}
\label{sec:methods}
Building on prior work, we adopt a UNet architecture \cite{ronneberger_unet_2015} with a custom MobileNet-v3-large encoder \cite{mobilenet_large} pretrained on ImageNet \cite{imagenet}. This configuration was previously found to provide the best trade-off between segmentation performance and computational efficiency for thermal cloud detection on FOREST-2 imagery by \citet{woelkiCloud}. Its lightweight design makes it particularly well-suited for deployment on resource-constrained hardware such as the NVIDIA Jetson Nano, which is used for onboard processing on FOREST-2 satellites.
To preserve the integrity of the pretrained encoder, we freeze its weights during training and update only the decoder. The model is trained using binary cross-entropy loss and optimized with the Adam optimizer. The models take as input only the LWIR2 band from FOREST-2 and the domain-adapted Band 6 from Landsat-7, normalized using Z-score normalization based on the mean and standard deviation of the respective dataset.
\subsection{Transfer Learning Strategies}
To evaluate the impact of using publicly available labeled cloud masking datasets, we define three inductive transfer learning setups:
\begin{itemize}
\item \textbf{Intra-Dataset Training and Testing (l7-l7):} As a baseline, we train and evaluate the model on the Landsat-7 dataset alone. This setup demonstrates the segmentation performance that can be achieved on a large, high-quality dataset using our chosen architecture. It serves as a reference point for assessing cross-dataset performance and the benefit of joint training.
\item \textbf{Cross-Dataset Generalization (l7-f2):} In this setting, the model is trained only on the Landsat-7 dataset and evaluated on the FOREST-2 dataset. This configuration tests the zero-shot generalization capabilities.
\item \textbf{Joint Training (joint-f2):} In this scenario, we train the model using the Landsat-7 training dataset with each fold of the FOREST-2 training data, and evaluate the model on the corresponding FOREST-2 test fold. This experiment shows how combining datasets from different sources can boost model performance compared to training solely on FOREST-2 data. Our approach follows a similar strategy to that described in \citet{MateoGarca2020}, where joint training and domain adaptation improved robustness and accuracy.
\end{itemize}
\subsection{Evaluation Metrics}
\label{subsec:metrics}
We report segmentation performance using the macro-averaged F1 score and overall accuracy. Accuracy reflects the overall proportion of correctly classified pixels and is commonly indicated in the literature on cloud segmentation. Our dataset is slightly imbalanced, with roughly two-thirds clear pixels, so we report the macro-averaged F1 score, which averages the per-class F1 scores equally. This metric provides a more balanced assessment than accuracy, which can be biased toward the dominant class.

\section{Results}
\label{sec:results}
We present the results of the three conducted experiments and the model's performance that was trained and tested with FOREST-2 data only (f2-f2). 

\begin{table}[ht]
\caption{Classification accuracies and Macro F1 scores for transfer learning experiments.}
\label{tab:results}
\vskip 0.15in
\begin{center}
\begin{small}
\begin{sc}
\begin{tabular}{lcccr}
\toprule
Experiment Name & Macro F1 & Accuracy\\
\midrule
MobileNet f2-f2    & 0.850    & 0.889\\
MobileNet joint-f2 & \textbf{0.877}    & \textbf{0.910}\\
MobileNet l7-f2    & 0.796    & 0.830\\
\textcolor{gray!80}{MobileNet l7--l7}    & \textcolor{gray!80}{0.903}    & \textcolor{gray!80}{0.917}\\
\bottomrule
\end{tabular}
\end{sc}
\end{small}
\end{center}
\vskip -0.1in
\end{table} 
\begin{figure}[ht]
\vskip 0.2in
\begin{center}
\centerline{\includegraphics[width=\columnwidth]{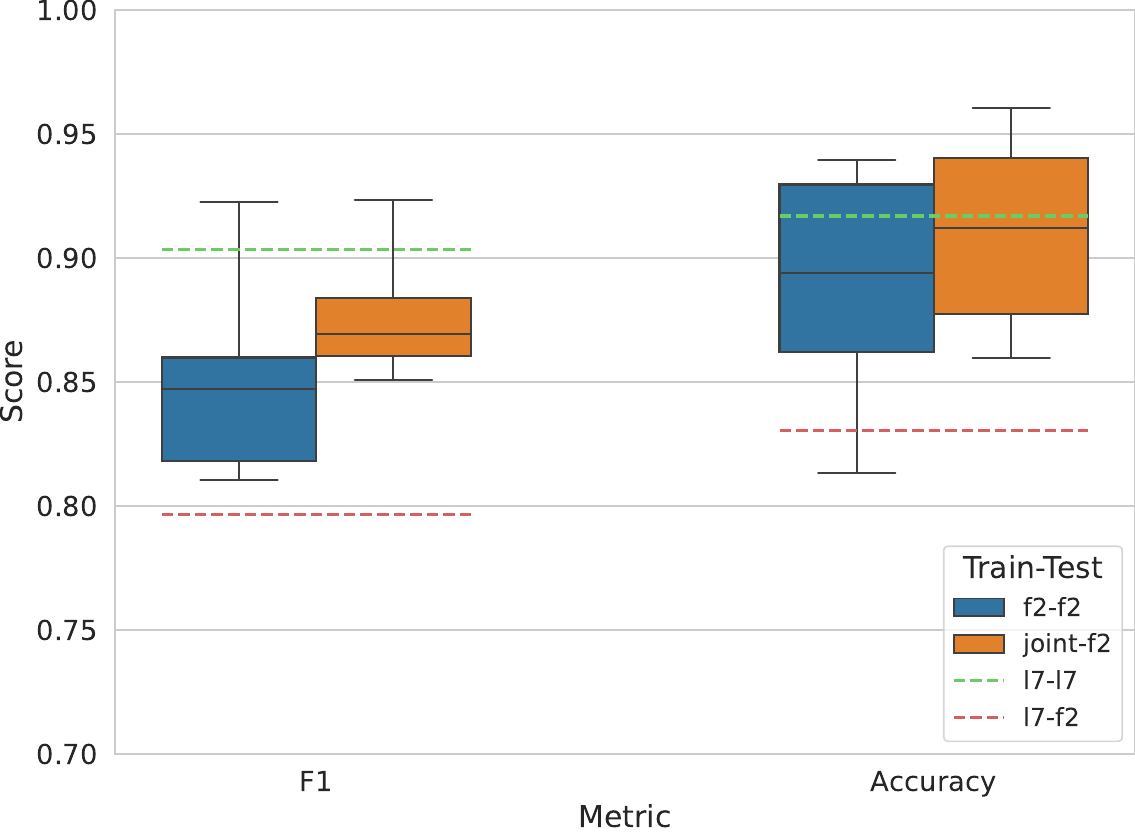}}
\caption{Macro F1 and accuracy across six folds for models trained on FOREST-2 only (f2-f2) and jointly with Landsat-7 (joint-f2). Joint training improves macro F1 from 0.850 to 0.877 and accuracy from 0.889 to 0.910. Dashed lines show Landsat-only results: l7-l7 (green) and l7-f2 (red).}
\label{fig:boxplot}
\end{center}
\vskip -0.2in
\end{figure}

The performance is shown in \autoref{fig:boxplot} and summarized in \autoref{tab:results}. While we evaluate the f2-f2 and joint-f2 experiments over all six folds, we test the models trained on only Landsat data on the Landsat test set (l7-l7) and all FOREST-2 data (l7-f2).

In the l7-l7 experiment, indicated in a lighter grey, the model achieves an accuracy of 91.7\% and a macro F1 score of 90.3\%. Overall, this is higher than the performance achieved when training and evaluating on FOREST-2 data (f2-f2), which achieves a macro F1 of 85\% and 88.9\% accuracy. This setup is a reference point for what we could expect from FOREST-2 with a similarly mature and comprehensive dataset.
When we apply the same model to the FOREST-2 dataset without fine-tuning (l7-f2), it still reaches 79.6\% F1 and 83\% accuracy, indicating good generalization despite the cross-domain shift. However, this still falls short of the performance achieved by the small, in-domain f2-f2 baseline.
The results from these experiments suggest that training on the larger and more diverse Landsat-7 dataset can enhance model performance in the FOREST-2 domain.
This is confirmed by the final experiment, where we jointly train FOREST-2 and Landsat-7 data (joint-f2), resulting in an F1 score of 87.7\% and an accuracy of 91\%. Compared to the f2-f2 baseline, this reflects an improvement of 2.7\% F1 and 2.1\% accuracy. In addition, the variability across the folds is reduced in terms of F1, suggesting improved generalization from the increased diversity of the training data, consistent with the findings of \citet{MateoGarca2020}.

The Precision-Recall and ROC curves (\autoref{fig:pr_roc}) and qualitative examples (\autoref{fig:examples}), both provided in \autoref{appendix}, show that the joint-f2 model achieves the best overall performance, reaching an AP of 0.92 and AUC of 0.96, compared to 0.90 and 0.94 for the f2-f2 baseline. It also performs better in visually complex regions and more reliably detects thin clouds, likely due to their higher representation in the Landsat-7 training data.

\begin{figure}[ht]
\vskip 0.2in
\begin{center}
\centerline{\includegraphics[width=\columnwidth]{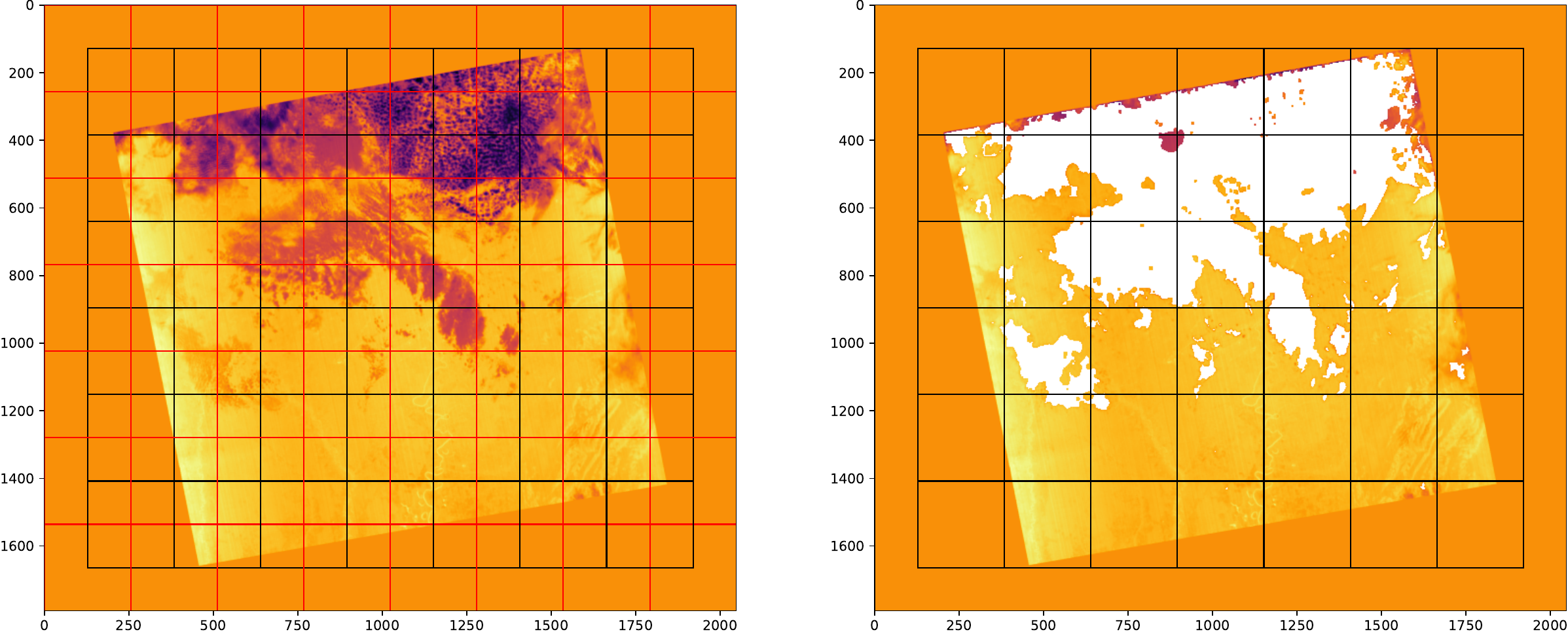}}
\caption{Onboard image tiling and stitching scheme. Red tiles are 512x512px, black tiles are 256x256px. The resulting cloud mask is shown in white.}
\label{fig:tiling}
\end{center}
\vskip -0.2in
\end{figure}

To enable real-time onboard processing, we deploy the joint-trained MobileNet model on an NVIDIA Jetson Nano with 4 GB RAM, representative of the hardware used on FOREST-2 satellites. Due to memory constraints, full-image inference is not feasible. Instead, we adopt a tiling strategy, which is visualized in \autoref{fig:tiling}: the input image is divided into overlapping 512×512 pixel tiles, each processed independently, and only the central 256×256 region of each tile is used to assemble the final prediction. This approach minimizes edge artifacts and ensures smooth transitions between tiles, resulting in a seamless cloud mask across the image.

We convert the trained PyTorch model to ONNX format to minimize dependencies and optimize runtime. The ONNX model is then compiled into a TensorRT engine, allowing high-throughput, low-latency inference on CUDA-enabled embedded devices like the NVIDIA Jetson. This optimization is crucial for our missions, where computational and memory resources are tightly constrained. On a full image (2691×1762 px), the final model achieves inference in under 5 seconds, with segmentation performance within 1\% of our on-ground implementation, which processes the whole image in one pass. These results confirm that accurate, thermal-only cloud masking can be performed efficiently onboard, supporting real-time decision-making and bandwidth-aware data prioritization in orbit.

\section{Discussion}
\label{sec:discussion}
Our study demonstrates that transfer learning is an effective strategy for thermal cloud segmentation, even when using single-band input. The l7-l7 setup provides a reference point, showing that mature, well-labeled, and globally representative datasets enable strong in-domain performance.

The l7-f2 experiment reveals promising zero-shot generalization: despite differences in platform design and data characteristics, the model achieves reasonable performance. This suggests thermal cloud patterns generalize across sensors, though the performance gap highlights residual domain shift. Narrowing this gap may be possible with improved domain adaptation techniques such as histogram matching.

The joint-f2 experiment confirms the benefit of combining public datasets with limited mission-specific data. With only 10\% of all training samples being from FOREST-2, this configuration surpasses the FOREST-2-only baseline and is approaching the performance of the Landsat-only model in its native domain. This further incentivises the development of a larger, globally representative FOREST-2 dataset, combined with more sophisticated domain adaptation techniques, which likely enhances the model's robustness to varying environmental conditions and cloud types. The decreased performance variability across the folds further supports the improved generalization ability.

The transfer learning approach has practical implications: it reduces labeling effort and supports satellite-specific fine-tuning via calibration using relatively small labeled subsets. Prelaunch training further enables cloud masking from the first day of operation.

In addition to its accuracy benefits, our model architecture supports efficient deployment in resource-constrained environments. Onboard inference on a Jetson Nano, representative of FOREST-2 hardware, runs in under five seconds for a full image using a tiling strategy and TensorRT optimization. This confirms the feasibility of thermal cloud masking on our CubeSat directly in orbit, enabling data prioritization and pre-filtering before downlink. Onboard capabilities are especially valuable for missions with limited bandwidth and power budgets.

While our results are encouraging, they are constrained by the limited size and geographic diversity of the FOREST-2 dataset. Future work should expand validation efforts across diverse conditions to better assess robustness. Moreover, domain adaptation techniques, such as SRF/point spread function (PSF) matching or adversarial training, could help bridge the remaining domain gap. Looking ahead, we expect that incorporating professionally labeled FOREST-2 data will further improve model accuracy and consistency. Finally, future work could explore whether combining LWIR2 with additional bands or indices enhances performance.

\section{Conclusion}
\label{sec:conclusion}
This work explored transfer learning for thermal cloud segmentation on CubeSat imagery using only a single thermal band. By leveraging a large, high-quality, publicly available, pre-labeled dataset from Landsat-7, we demonstrated that pretraining on mature imagery and fine-tuning on a small set of mission-specific samples can significantly improve segmentation performance for emerging CubeSat missions, even on a single thermal band. Our joint training approach improved the macro F1 score from 0.850 to 0.877 and yielded more consistent predictions across folds, confirming the benefits of training on diverse data sources.

Despite relying solely on thermal input, our models reach accuracy levels comparable to mature cloud masking systems that use multispectral data and auxiliary features, staying within the computational limits of a CubeSat. This highlights the potential of lightweight, thermal-only deep learning approaches, especially in resource-constrained thermal CubeSat environments. At the same time, our evaluation remains limited compared to operational benchmarks, which are validated across diverse 
conditions, including snow-covered terrains, coastal regions, and various biomes, to understand their specific limitations. Refining our validation set and incorporating professionally labeled data will be essential for robust benchmarking in future work.

Our results suggest that satellite-specific fine-tuning with minimal labeled data after leveraging pre-labeled datasets is a viable option to create an efficient and accurate cloud segmentation model for thermal-only onboard processing on a CubeSat. 

\section*{Acknowledgements}
The authors would like to thank the anonymous reviewers for their helpful feedback and for highlighting several important directions for future work. We also thank our colleagues at OroraTech for valuable discussions and support throughout the development of this work.

\section*{Impact Statement}
This paper addresses cloud detection on CubeSat missions using thermal data to improve preprocessing for remote sensing applications under constrained hardware and data availability. The resulting methods support humanitarian use cases such as wildfire detection, drought monitoring, and urban heat mapping, particularly in regions where low-cost CubeSat constellations enable frequent thermal imaging at a global scale. This work may help reduce latency and labeling effort in early-stage EO missions by leveraging public datasets and promoting prelaunch transfer learning.

As with many satellite data applications, the techniques developed here could, with modification, be applied to dual-use scenarios, including military or intelligence-gathering satellites. However, this work is motivated by civilian and environmental monitoring goals, and we do not specifically tailor our methods for non-civilian use. We encourage future deployments to ensure transparency and alignment with international norms for peaceful Earth observation.

\bibliography{bibliography}
\bibliographystyle{icml2025}

\newpage
\appendix
\onecolumn
\section{Appendix}
\label{appendix}
\begin{figure}[ht!]
\begin{center}
\centerline{\includegraphics[width=\columnwidth]{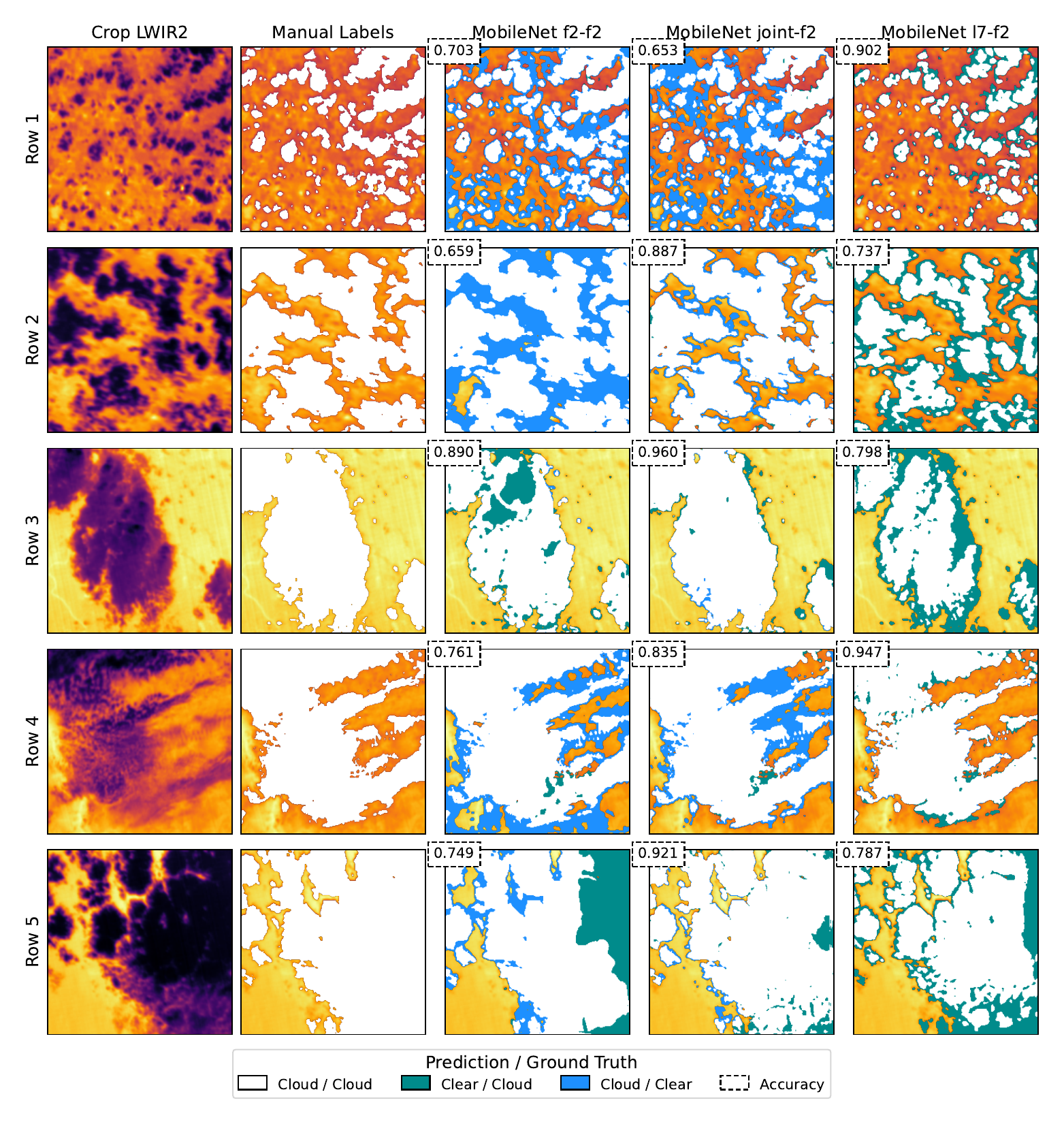}}
\caption{Qualitative comparison of cloud segmentation results on five representative FOREST-2 crops. Each row shows the original thermal input, the corresponding manual annotation, and the predicted masks from the evaluated models. The joint-trained model performs particularly well in visually complex regions and appears to handle thin cloud structures more accurately, likely due to the greater presence of such cases in the Landsat-7 training data. Notably, the l7-f2 model also produces reasonable segmentation results, despite having never seen FOREST-2 data during training.}
\label{fig:examples}
\end{center}
\end{figure}

\begin{figure}[ht]
\vskip 0.2in
\begin{center}
\centerline{\includegraphics[width=\columnwidth]{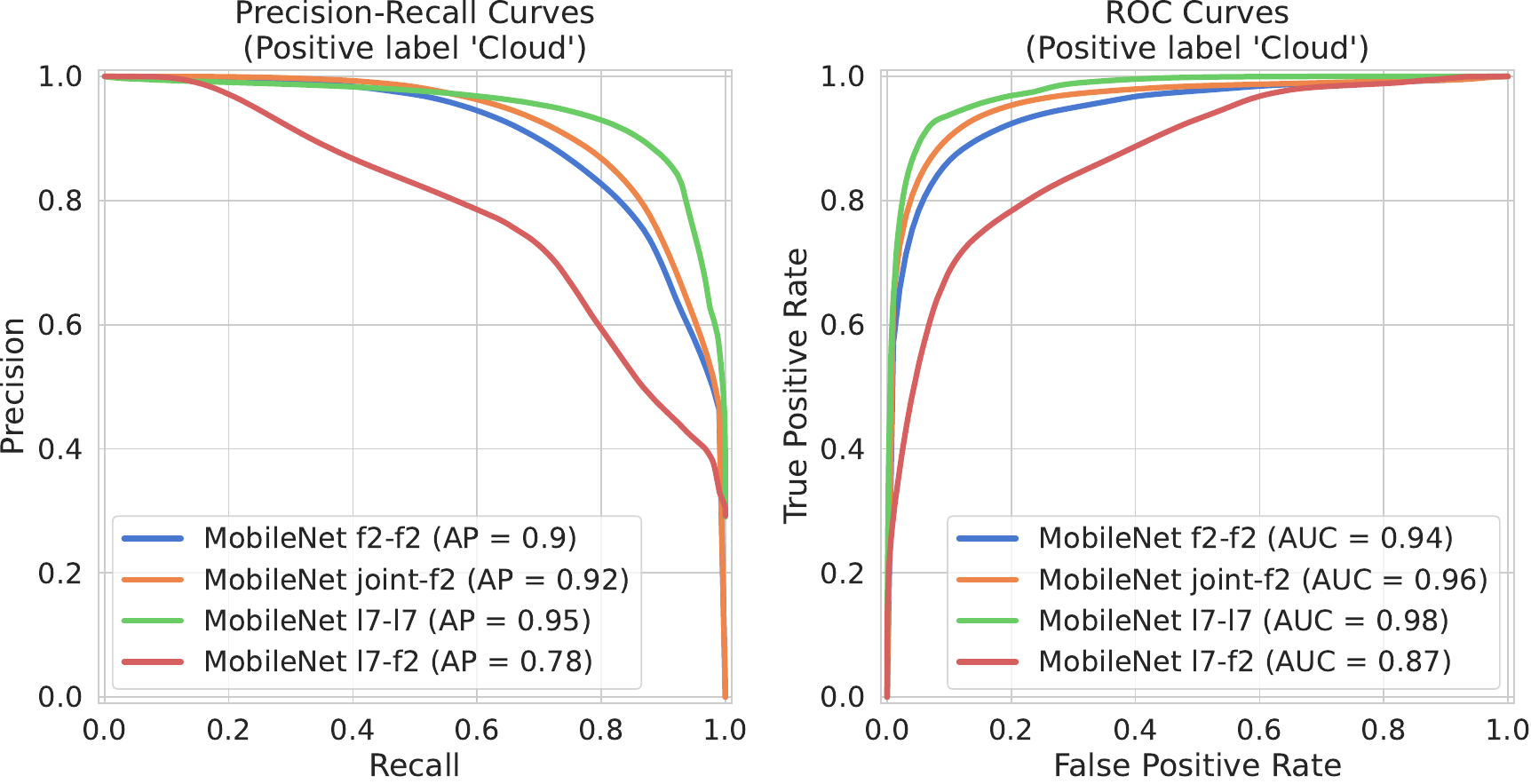}}
\caption{Receiver Operating Characteristic and Precision-Recall curves for all models. Joint training (joint-f2) achieves an AP of 0.92 and AUC of 0.96, outperforming the f2-f2 baseline (AP = 0.90, AUC = 0.94). The l7-f2 model shows lower generalization performance (AP = 0.78, AUC = 0.87), while l7-l7 reaches the highest scores overall (AP = 0.95, AUC = 0.98).}
\label{fig:pr_roc}
\end{center}
\vskip -0.2in
\end{figure}

\end{document}